\documentclass[10pt,twocolumn,letterpaper]{article}

\usepackage{cvpr}              

\usepackage{times}
\usepackage{epsfig}
\usepackage{graphicx}
\usepackage{amsmath,amssymb}
\usepackage{booktabs}
\usepackage{multirow}
\usepackage{xcolor}
\usepackage[section]{placeins}
\usepackage{svg}
\usepackage{subcaption}
\usepackage{url}
\usepackage{adjustbox}
\usepackage{pifont}
\usepackage{comment}
\newcommand{\cmark}{\ding{51}}
\newcommand{\xmark}{\ding{55}}

\definecolor{cvprblue}{rgb}{0.21,0.49,0.74}
\usepackage[pagebackref,breaklinks,colorlinks,allcolors=cvprblue]{hyperref}

\def\paperID{CVsports-33} 
\def\confName{CVPR}
\def\confYear{2026}

\title{\textbf{GRAZE}: \textbf{G}rounded \textbf{R}efinement and 
Motion-\textbf{A}ware \textbf{Z}ero-Shot \textbf{E}vent Localization}

\author{
Syed Ahsan Masud Zaidi$^{1}$ \quad
Lior Shamir$^{1}$ \quad
William Hsu$^{1}$ \quad
Scott Dietrich$^{2}$ \quad
Talha Zaidi$^{1}$\\[0.5em]
$^{1}$Kansas State University, Manhattan, KS, USA\\
$^{2}$Albright College, Reading, PA, USA\\[0.25em]
{\tt\small ahsanzaidi@ksu.edu, lshamir@ksu.edu, bhsu@ksu.edu}\\
{\tt\small sdietrich@albright.edu, tzaidi@ksu.edu}
}

\begin{document}
\maketitle


\begin{abstract}
American football practice generates video at scale, yet the interaction of interest occupies only a brief window of each long, untrimmed clip. Reliable biomechanical analysis, therefore, depends on spatiotemporal localization that identifies both the interacting entities and the onset of contact. We study First Point of Contact (FPOC), defined as the first frame in which a player physically touches a tackle dummy, in unconstrained practice footage with camera motion, clutter, multiple similarly equipped athletes, and rapid pose changes around impact.
We present GRAZE, a training-free pipeline for FPOC localization that requires no labeled tackle-contact examples. GRAZE uses Grounding DINO to discover candidate player-dummy interactions, refines them with motion-aware temporal reasoning, and uses SAM2 as an explicit pixel-level verifier of contact rather than relying on detection confidence alone. This separation between candidate discovery and contact confirmation makes the approach robust to cluttered scenes and unstable grounding near impact.

On $738$ tackle-practice videos, GRAZE produces valid outputs for $97.4\%$ of clips and localizes FPOC within $\pm 10$ frames on $77.5\%$ of all clips and within $\pm 20$ frames on $82.7\%$ of all clips. These results show that frame-accurate contact onset localization in real-world practice footage is feasible without task-specific training.
\end{abstract}
\section{Introduction}

Practice video recordings in contact sports like American football 
provide useful information on player technique and can help improve 
the performance of early-career athletes. However, the unstructured 
nature of raw footage remains a barrier to automated analysis. Raw 
practice clips are long, multi-scene, and contain both spatial and 
temporal noise, mixing warm-up drills, recovery periods, and coaching 
instructions around only a few seconds of biomechanically relevant 
action. Coaches, athletic trainers, and biomechanists require that 
action window indexed accurately and reproducibly. While 
spatio-temporal datasets such as AVA~\cite{gu2018avavideodatasetspatiotemporally} 
have formalized atomic action localization in broadcast footage, they 
provide no mechanism for contact onset detection in unconstrained 
monocular recordings.

In this paper, we study \emph{First Point of Contact} (FPOC) 
localization in tackle-practice video, defined as the first frame 
at which a player makes physical contact with a padded tackle dummy. 
FPOC is the natural temporal anchor for contact-centered analysis: 
strike-zone assessment, postural alignment at impact, and kinematic 
profiling all require knowing exactly when contact began. An incorrect FPOC shifts pose measurements to pre- or post-contact postures, making rubric-based scoring unreliable. Solving this requires both identifying the correct player-dummy pair in a cluttered scene and determining  the precise frame of contact.

\begin{figure}
    \centering
    \includegraphics[width=1\linewidth]{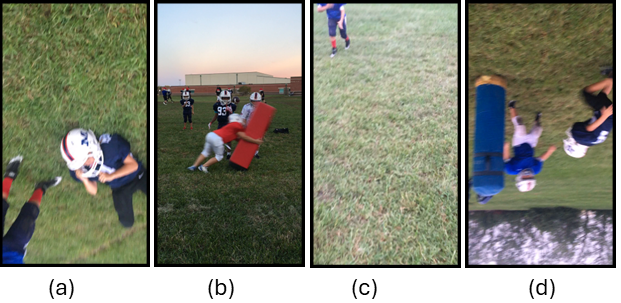}
    \caption{Representative recording challenges in tackle-practice videos.
    \textbf{(a)} Variable scene composition with multiple players, no visible tackle dummy, and a non-sprinting actor under unstable camera motion.
    \textbf{(b)} Multiple players with mutual occlusion near the contact window.
    \textbf{(c)} Player and dummy partially or fully absent due to 
    abrupt panning.
    \textbf{(d)} Inverted camera orientation with motion blur and 
    multi-player overlap onto the dummy.}
    \label{fig:challenges}
\end{figure}

Achieving this from raw practice footage is hard for reasons standard sports-video benchmarks do not capture. Cameras are handheld or field-mounted, panning and jittering throughout. Backgrounds shift abruptly between repetitions. Multiple athletes in similar gear share the frame, and some stand between the camera and the active player, creating distractors that score highly under any appearance-based detector. Standard bounding boxes fail in both directions: they can overlap the dummy without physical contact, and at the true moment of contact, occlusion can eliminate that overlap entirely. Representative failure conditions are shown in Figure~\ref{fig:challenges}.

The fundamental difficulty is the gap between \emph{pair discovery} 
and \emph{contact confirmation}. Detection confidence reflects 
appearance similarity to a text prompt, not physical intersection. 
What FPOC localization requires is pixel-level evidence that the two 
objects overlap at the earliest frame where that overlap occurs, and 
a strategy that stays robust when grounding is unstable across 
consecutive frames.

We address FPOC localization with a training-free pipeline requiring 
no task-specific fine-tuning and no labeled examples from this domain. 
The central insight is to treat SAM2~\cite{ravi2024sam2} not as a 
passive segmentation backend but as an explicit \emph{pixel-level 
contact verifier}: given candidate player-dummy pairs, we propagate 
masks with SAM2 and use their intersection as direct contact evidence, 
decoupled from grounding confidence. This enables a ranked 
multi-candidate fallback in which all grounding candidates across 
prompt sets, temporal positions, and detection thresholds are 
evaluated until SAM2 mask overlap independently confirms contact. 
Candidates are generated using Grounding 
DINO~\cite{liu2023groundingdino} with hierarchical prompt scheduling 
and progressive threshold relaxation, then ranked by a geometric 
motion signal combining displacement magnitude with cosine similarity 
between the player's motion vector and the approach vector toward 
the dummy. Because grounding succeeds most reliably at mid-contact, 
when both objects are simultaneously salient, the grounding frame 
is systematically biased later than the true onset. A two-phase 
backward refinement using sequential backtracking and binary search 
corrects this bias and recovers the earliest consistent frame.

This paper makes three contributions. First, we introduce a 
training-free spatiotemporal localization pipeline for FPOC detection 
in unconstrained tackle-practice footage, to our knowledge the first 
system targeting frame-accurate contact onset localization in this 
domain at scale. Second, we propose using SAM2 mask propagation as 
an independent pixel-level contact verifier, decoupling detection 
confidence from contact evidence and enabling a multi-candidate 
fallback that persists until mask overlap confirms contact. Third, 
we introduce a geometric directional motion scoring strategy based 
on cosine similarity between the player's motion vector and the 
dummy approach vector, combined with displacement magnitude, to 
stabilize candidate ranking without any learned classifier or tracker.

\subsection{Related Work}

Sports video analysis has advanced rapidly, but largely under 
conditions that practice footage does not satisfy. Broadcast cameras 
track the action deliberately with controlled lighting, and benchmarks 
like SoccerNet~\cite{Giancola_soccernet,deliege2021soccernetv2} and 
action spotting formulations~\cite{cioppa2020context} have pushed 
temporal precision toward single-frame accuracy in that setting. 
Strong backbones such as I3D~\cite{carreira2018} and 
SlowFast~\cite{feichtenhofer2019slowfast} were designed and evaluated 
on curated datasets that assume a clean, actor-centered field of view, and survey work has traced how heavily broadcast-era gains depend on those assumptions~\cite{wu2022surveyvideoactionrecognition}. American 
football practice looks almost nothing like this. Formation recognition and play-tagging methods exist for the sport~\cite{Newman2021,electronics12030726,nayab2025}, and more recent work has begun treating tackle practice as a domain in its own right~\cite{zaidi2025vits}, but these contributions target clip-level labels and strategic structure, leaving frame-accurate contact onset unaddressed.

Tackle safety has attracted a parallel line of work. Nafi et al.~\cite{Nafi2022RiskyTackle3D} classify clips by risk level using 3D convolutional networks, demonstrating that a useful injury screening signal exists in practice footage, but clip-level labels do not identify when contact begins. A follow-on effort~\cite{nafiSegmentation} uses Cascade R-CNN to segment individual players against cluttered backgrounds. Image-level segmentation is temporally inert, however: it provides no mechanism for maintaining identity across frames or for deciding when two actors first intersect. Spatial and temporal problems are handled separately throughout this body of work; our setting requires both to be solved together.

Because visual appearance varies substantially across sessions and 
equipment configurations, training a task-specific detector is 
impractical. Contrastive image, language pretraining~\cite{radford2021clip} demonstrated that aligning vision 
and language at scale yields representations that transfer to unseen 
categories, motivating a generation of open-vocabulary detectors. 
GLIP~\cite{li2022glip} unified grounding and detection through 
language-image pretraining; OWL-ViT~\cite{minderer2022owlvit} showed 
that a contrastively trained Vision Transformer achieves competitive 
open-set detection without task-specific heads; and Grounding 
DINO~\cite{liu2023groundingdino} fused a DINO-style detector with 
cross-modal attention for strong open-vocabulary performance. 
Phrase-level grounding is particularly suited here because detailed 
descriptions of player posture and equipment help separate the active 
tackler from bystanders. The limitation shared by all these methods 
is that per-frame confidence measures co-occurrence, not contact.

Video segmentation can close this gap if used correctly. 
Self-supervised pretraining~\cite{tong2022videomae} has shown that 
temporal masked autoencoding yields representations that generalize 
broadly, and SAM~\cite{kirillov2023SAM1} established promptable 
segmentation at image scale. SAM2~\cite{ravi2024sam2} extends this to 
video via a streaming memory architecture. We depart from the standard 
use pattern, where SAM2 is a backend that returns masks for downstream 
use, by treating mask intersection itself as a contact detection 
signal. When a propagated player mask begins to overlap a separately 
propagated dummy mask, that is direct geometric evidence of contact, 
wholly independent of detection confidence.

Supervised temporal action localization methods such as BMN~\cite{lin2019bmn} and ActionFormer~\cite{zhang2022actionformer} 
achieve strong results on curated benchmarks but require frame-level 
annotations unavailable for contact onset in practice footage. 
The closest annotation-free relatives are zero-shot temporal methods 
T3AL~\cite{liberatori2024t3al} and ZEETAD~\cite{phan2024zeetad}, and 
natural-language moment retrieval~\cite{gao2017tall}. All produce 
temporal segments associated with action categorie, a useful output 
for recognition but the wrong granularity for contact onset. At 30 fps, a half-second segment window spans 15 frames, and whether a measurement lands on pre-contact posture or mid-contact deformation depends on which of those frames is selected. More fundamentally, these methods respond to what an action looks like; FPOC is determined by whether two specific objects physically intersect, a question no 
appearance-based segment predictor is designed to answer.

\section{Methodology}

\subsection{Problem Formulation}

Let $\mathcal{V} = \{I_t\}_{t=1}^{T}$ be an untrimmed video 
of $T$ frames. The scene contains at least one player $P$ 
approaching a stationary tackle dummy $D$. Given $\mathcal{V}$, 
we seek two temporal quantities: the First Frame with Both 
Objects ($t_{\text{FFBO}}$), the earliest frame in which both 
$P$ and $D$ are simultaneously visible, and the First Point of 
Contact ($t_{\text{FPOC}}$), the exact frame at which the player first physically intersects the dummy. We also produce per frame pixel masks $\mathcal{M}^{(P)}_t$ and $\mathcal{M}^{(D)}_t$ over the event window $[t_{\text{FFBO}}, t_{\text{end}}]$, where $t_{\text{end}}$ is defined in Eq.~\ref{eq:window_end}. 
The pipeline uses no task-specific fine-tuning and no labeled 
examples from this domain.

\subsection{Pipeline Overview}

\begin{figure*}[t]
    \centering
    \includegraphics[width=0.95\linewidth]
    {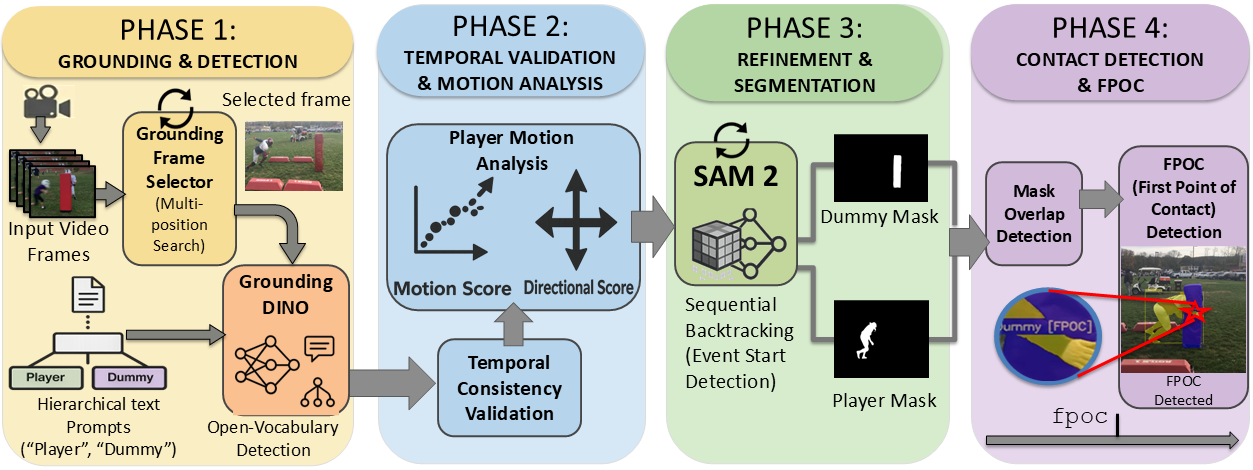}
    \caption{Four-phase pipeline. \textbf{(1) Grounding:} 
    Grounding DINO searches multiple temporal positions with 
    hierarchical prompts and progressive threshold relaxation, 
    retaining all valid player-dummy candidates. 
    \textbf{(2) Validation:} Each candidate is scored by 
    temporal consistency, displacement magnitude, and directional 
    approach toward the dummy; candidates are ranked before SAM2 
    evaluation. \textbf{(3) Refinement and Segmentation:} 
    Backward refinement recovers $t_{\text{FFBO}}$; SAM2 
    propagates player and dummy masks from that frame. 
    \textbf{(4) Contact Verification:} FPOC is the first frame 
    where the propagated masks overlap. If no overlap is found, 
    the next ranked candidate is evaluated.}
    \label{fig:pipeline}
\end{figure*}

Figure~\ref{fig:pipeline} summarizes the four-phase approach. Grounding proposes candidate player-dummy bounding-box pairs, which a motion analysis module ranks by geometric plausibility. Backward refinement then corrects the temporal bias that arises when grounding latches onto a mid-contact frame rather than the true event onset. SAM2 verifies contact at the pixel level. Because grounding confidence and physical contact are distinct quantities, we treat them separately, i.e., a candidate is accepted only when its propagated masks produce measurable pixel overlap, regardless of how confidently it was detected.

\subsection{Hierarchical Grounding and Candidate Generation}

Player appearance varies across practice sessions depending on attire, equipment, and lighting conditions. A single prompt cannot cover this range reliably. We use a three-level hierarchy $\mathcal{P} = \{P_{\text{gear}}, P_{\text{nogear}}, P_{\text{generic}}\}$, ordered from most to least specific. For instance, $P_{\text{gear}}$ describes a helmeted player sprinting in a forward lean, while $P_{\text{generic}}$ describes only a person running toward a red object. The dummy prompt is fixed across all three levels, as the tackle dummy has a consistent physical description. Each player-dummy prompt pair is submitted jointly to Grounding DINO~\cite{liu2023groundingdino} as a single phrase-grounding query, grounding both noun phrases in one forward pass.

\begin{figure}[!htbp]
    \centering
    \includegraphics[width=\linewidth]{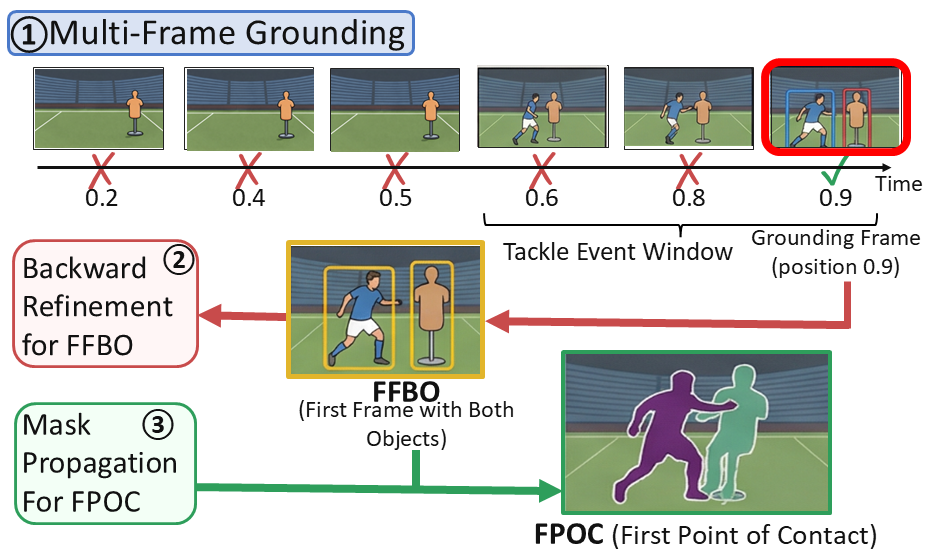}
    \caption{Temporal localization of first contact using multiframe grounding and backward refinement. 
    \textbf{(1)} Multi-frame grounding queries the untrimmed clip at a set of normalized time positions; in this example, a valid player-dummy pair is first detected at position 0.9, defining a tackle event window.
    \textbf{(2)} Backward refinement then steps backward from the grounding frame to find FFBO (the first frame in which both the player and dummy are simultaneously present).
    \textbf{(3)} Starting from FFBO, SAM2 propagates masks forward and identifies FPOC as the earliest frame where the propagated masks provide evidence of player-dummy contact.}
    \label{fig:temporal}
\end{figure}

We probe six temporal positions across the video, searching a local offset window around each at three progressively relaxed box-confidence thresholds. Rather than returning on the first success, we collect every valid candidate across all positions, thresholds, and prompt levels. Figure~\ref{fig:temporal} illustrates this search. This exhaustive collection matters in practice because grounding quality and contact quality do not correlate monotonically; a strong detection response at mid-contact may track the wrong player, while a weaker detection at an earlier frame can still confirm the correct onset through mask intersection.

\textbf{Detection classification.}
Raw detections $\{d_i\}$ carry bounding boxes $b_i \in \mathbb{R}^4$, confidence scores $s_i$, and phrase labels $l_i$. Each detection is assigned to the player or dummy class based on label keywords and geometry. Let $\mathcal{K}_d = \{\textit{dummy, pad, bag, tackle, training, blocking}\}$ and $\mathcal{K}_p = \{\textit{player, person, athlete, helmet, football, running,  sprint}\}$. Let $\text{AR}(d_i) = h_i / w_i$ denote the height-to-width aspect ratio, $\tau_{\text{area}} = 0.01 \cdot HW$ a minimum area threshold, and $\text{edge}(d_i)$ an indicator that the detection center lies within 15\% of the frame width from either horizontal edge or within $10\%$ from the top:

\begin{equation}
\text{kind}(d_i) = \begin{cases}
\text{dummy} & \text{if } (\text{AR}(d_i) > 2.0) 
               \vee (l_i \in \mathcal{K}_d) \\[4pt]
\text{player} & \text{if } l_i \in \mathcal{K}_p \\
              & \phantom{\text{if }} 
                \wedge\ \text{area}(d_i) > \tau_{\text{area}} \\
              & \phantom{\text{if }} 
                \wedge\ \neg \text{edge}(d_i)
\end{cases}
\label{eq:classify}
\end{equation}

where $edge(d_i)$ is true when the detection center lies within the frame boundary margins. Dummies with $\text{AR}(d_i) < 0.8$ are rejected as horizontal objects, since a practice dummy is always upright. Players are not filtered by aspect ratio: a player mid-tackle may be nearly horizontal. Among valid dummy detections, the best candidate $D^*$ is selected by a weighted score over confidence $s_d$, normalized distance $\text{dist}_{\text{c}}(d)$ from the image center, and verticality:

\begin{equation}
\begin{aligned}
D^* = \arg\max_{d \in \mathcal{D}_{\text{dummy}}} \Big[
&\; 0.4\, s_d 
+ 0.3\bigl(1 - \text{dist}_{\text{c}}(d)\bigr) \\
&+ 0.3\,\min\!\bigl(\tfrac{\text{AR}(d)}{3.0},\, 1\bigr)
\Big]
\end{aligned}
\label{eq:dummy_select}
\end{equation}

The top three player candidates ranked by proximity to $D^*$ 
are forwarded to the validation stage.

\subsection{Temporal Validation and Motion Scoring}

A detection at a single frame can be a false positive arising 
from motion blur, partial occlusion, or a background athlete 
passing near the dummy. Each candidate at primary frame $t_0$ 
is matched against detections in 14 neighboring frames 
$\mathcal{F}_{\text{val}} = \{t_0 + \delta\}$ via a composite 
confidence score over normalized IoU, center displacement, and 
area similarity. Let $\tau_{\text{IoU}} = 0.20$ and 
$\tau_{\text{dist}} = 200$ px. The size ratio is 
$\text{sr}(d_0, d_v) = \min(A_0,A_v) / \max(A_0,A_v)$, where 
$A$ denotes box area:

\begin{equation}
\begin{aligned}
\text{conf}(d_0, d_v) = 
&\; 0.55\,\tfrac{\text{IoU}(d_0,d_v)}{\tau_{\text{IoU}}}
+ 0.30\Bigl(1 - \tfrac{\text{dist}(d_0,d_v)}{\tau_{\text{dist}}}\Bigr) \\
&+ 0.15\;\text{sr}(d_0, d_v)
\end{aligned}
\label{eq:match_conf}
\end{equation}

The mean of $\text{conf}(d_0, d_v)$ across all matched frames defines the consistency score $c_{\text{cons}}$. A player candidate must match in at least 3 of the 14 frames; a dummy candidate requires at least 2. Let $\mathcal{Q}$ 
denote the set of matched player detections across 
$\mathcal{F}_{\text{val}}$, with $c_m$ the center of match 
$m \in \mathcal{Q}$ and $c_0$ the center at $t_0$.

Temporal persistence alone does not separate the active 
tackler from a stationary bystander who appears consistently 
near the dummy. We therefore compute two geometric signals from the matched player centers across $\mathcal{F}_{\text{val}}$. The \textbf{displacement score} measures how much the player has moved across the validation window, normalized by a reference distance of 200 pixels:

\begin{equation}
m_{\text{disp}}(P) = \min\!\left(\frac{1}{|\mathcal{Q}|}
\sum_{m \in \mathcal{Q}} 
\frac{\lVert c_0 - c_m \rVert_2}{200},\; 1\right)
\label{eq:motion}
\end{equation}

The \textbf{directional approach score} measures whether the 
observed motion points toward the dummy. Let 
$\bar{c}_{\text{past}}$ be the mean center of all matched 
detections with $\text{frame\_idx} < t_0$. The unit motion 
vector is $\hat{v}_{\text{motion}} = (c_0 - \bar{c}_{\text{past}}) 
/ \lVert c_0 - \bar{c}_{\text{past}} \rVert$ and the unit 
approach vector is $\hat{v}_{\text{to-dummy}} = (c_{D^*} - c_0) 
/ \lVert c_{D^*} - c_0 \rVert$, where $c_{D^*}$ is the dummy 
center. Their cosine similarity is scaled to $[0, 1]$:

\begin{equation}
m_{\text{dir}}(P,D) = 
\frac{\langle \hat{v}_{\text{motion}},\, 
\hat{v}_{\text{to-dummy}} \rangle + 1}{2}
\label{eq:directional}
\end{equation}

Candidates with $m_{\text{disp}} < 0.08$ or $m_{\text{dir}} 
< 0.30$ are discarded. The surviving candidates are ranked by:

\begin{equation}
\text{conf}_{\text{overall}} = 
0.3\,c_{\text{cons}} + 
0.3\,m_{\text{disp}} + 
0.4\,m_{\text{dir}}
\label{eq:overall}
\end{equation}

SAM2 evaluates candidates in descending order of 
$\text{conf}_{\text{overall}}$, stopping at the first that 
produces mask overlap.

\subsection{Two-Phase Backward Refinement}

Grounding is most reliable near mid-contact, when both objects 
are large and simultaneously salient. As a result, $t_g$ 
tends to fall later than the true event start $t_{\text{FFBO}}$. 
Two phases correct this. In \textbf{Phase~1}, the pipeline 
steps backward from $t_g$ one frame at a time, re-running 
detection with the reference boxes as spatial anchors and 
tolerating at most one consecutive miss before stopping. This 
recovers frames where both objects remain jointly detectable 
without over-extending into frames that precede the drill. 
In \textbf{Phase~2}, exponential offsets $\{5, 10, 20, 50\}$ 
frames back from the current boundary are probed. When both 
objects are found at a probe frame, binary search localizes 
the exact earliest consistent frame. The result is 
$t_{\text{FFBO}}$, and boxes re-extracted at that frame serve 
as the SAM2 seed prompts.

\subsection{SAM2 Contact Verification and FPOC}

SAM2~\cite{ravi2024sam2} is initialized at $t_{\text{FFBO}}$ with the refined bounding boxes and propagates separate binary masks for the player ($i{=}P$) and dummy ($i{=}D$) forward through the remaining frames:

\begin{equation}
\mathcal{M}^{(i)}_t = \text{SAM2}
\!\left(\mathcal{V},\, t_{\text{FFBO}},\, b^{(i)}\right),
\quad i \in \{P, D\}
\label{eq:sam2}
\end{equation}

where $\mathcal{M}^{(i)}_t \in \{0,1\}^{H \times W}$ is the 
binary mask for object $i$ at frame $t$, and $b^{(i)}$ is the 
corresponding seed box. Contact at each frame is quantified as 
the pixel-wise logical intersection of the two propagated masks:

\begin{equation}
\text{overlap}_t = \sum_{x,y} 
\mathcal{M}^{(P)}_t(x,y) \;\wedge\; \mathcal{M}^{(D)}_t(x,y)
\label{eq:overlap}
\end{equation}

The FPOC is the earliest frame where overlap reaches at least 
one pixel ($\tau_{\text{overlap}} = 1$):

\begin{equation}
t_{\text{FPOC}} = \min\bigl\{t \;\big|\; 
\text{overlap}_t \ge \tau_{\text{overlap}}\bigr\}
\label{eq:fpoc}
\end{equation}

If no such frame exists for the current candidate, it is 
rejected and the next in the ranked list is evaluated. This 
is the mechanism driving the multi-candidate fallback: contact 
evidence, not detection confidence, decides which candidate is 
accepted.

In untrimmed practice footage, there is no consistent visual marker that defines when a tackle ends. 
Some drills stop immediately after contact, while others continue as the player drives through contact and keeps moving with the dummy. 
Since our goal is onset localization, we avoid estimating an end time from mask overlap and instead output a compact, event-centered window that includes the approach and the initial collision.

We use a fixed post-contact tail. Let $L_{\text{post}}$ be the number of frames retained after contact. We define
\begin{equation}
t_{\text{end}} = \min\!\bigl(T,\; t_{\text{FPOC}} + L_{\text{post}}\bigr),
\label{eq:window_end}
\end{equation}
and output masks and frames over $\mathcal{W} = [\,t_{\text{FFBO}},\, t_{\text{end}}\,]$. 
Unless stated otherwise, we use $L_{\text{post}} = 20$ frames.

A result is accepted unless both $\text{conf}_{\text{overall}} < 0.25$ and $m_{\text{dir}} < 0.20$ hold simultaneously, which filters genuinely degenerate detections while preserving valid FPOC estimates with modest overall scores. 
Clips where no candidate confirms contact are flagged for manual review.


\section{Experiments}

\subsection{Dataset}

We evaluated on 738 untrimmed monocular videos recorded at multiple
American football training facilities using consumer-grade smartphones and fixed-mount cameras at 30\,fps. Recording conditions vary substantially in background, viewing angle, distance to the drill, and illumination. The player population spans youth athletes through adults; attire ranges from full football uniform to casual clothing. This variability in the dataset helps keep the dataset natural and generalizes to represent real-world practice footage. The Tackle dummies used in the dataset are upright rectangular padded targets. Clips were recorded in both indoor and outdoor settings, with the
majority outdoors. No clips were excluded for quality, and no domain-specific fine-tuning was applied to any component.

Ground-truth FPOC labels were obtained via frame-by-frame manual annotation: the annotator marked the first frame in which the player physically collides with and \emph{moves} the dummy (not merely the first frame of visual overlap/occlusion). We estimate an annotation uncertainty of approximately $\pm 3$ frames. Out of 738 videos, \textbf{681} include a verified ground-truth FPOC label; the remaining 57 were recorded without a simultaneous annotation pass. We define the \emph{evaluable subset} as the intersection between the ground-truth pool (681 clips) and the set of clips for which a given method successfully produces a valid segmentation/tracking output. Consequently, in the ablation study the evaluable subset size varies by setting (Table~\ref{tab:ablation}). Our full system segments 719 clips, yielding an evaluable subset of 666.

\subsection{Evaluation Protocol}

FPOC localization has an asymmetric cost structure: early predictions can be trimmed in post-processing, whereas late ones shift downstream pose measurements into post-contact postures.
We report tolerance-window accuracy: prediction $\hat{t}$ is correct if $|\hat{t} - t^{*}| \le \epsilon$ frames, for
$\epsilon \in \{5, 10, 15, 20\}$ frames. End-to-end accuracy fixes the denominator at all 738 videos, penalizing any clip the system fails to process. Conditional precision restricts the denominator to the evaluable subset, isolating per-clip localization quality from coverage.  Our ablation isolates the contribution of each pipeline component, grounding breadth, motion filtering, and backward refinement, all of which are specific to the contact-onset problem.

\subsection{Baselines}
We construct three internal ablation baselines by selectively enabling 
pipeline components (Table~\ref{tab:ablation_config}). All four 
configurations share identical SAM2 mask propagation and overlap-based 
FPOC detection. The ablation exclusively targets the grounding and 
candidate-selection stages upstream of SAM2, keeping the 
contact-verification step constant and ensuring clean attribution of 
performance differences.

\begin{table}[!htbp]
\centering
\caption{Ablation component configuration.}
\label{tab:ablation_config}
\setlength{\tabcolsep}{4pt}
\renewcommand{\arraystretch}{1.1}
\footnotesize
\resizebox{\columnwidth}{!}{%
\begin{tabular}{lcccc}
\toprule
Component
  & \textbf{SOLE} & \textbf{TRACE} & \textbf{MARS} & \textbf{GRAZE (ours)} \\
\midrule
Single prompt + fixed threshold   & \cmark & \cmark & \cmark & \xmark \\
Multi-prompt search               & \xmark & \xmark & \xmark & \cmark \\
Temporal validation               & \xmark & \cmark & \xmark & \cmark \\
Motion scoring                    & \xmark & \xmark & \cmark & \cmark \\
Backward refinement               & \xmark & \cmark & \xmark & \cmark \\
Multi-candidate retry             & \xmark & \xmark & \xmark & \cmark \\
\bottomrule
\end{tabular}%
}
\end{table}

Baseline \textbf{B1: }\textbf{SOLE} (Single-prompt Open-vocabulary Localization Estimator) 
uses a single fixed prompt and a fixed box-confidence threshold, with no temporal validation, motion scoring, or backward refinement. It 
represents the minimum viable grounding front-end feeding into the shared SAM2 contact verifier.

Baseline \textbf{B2: }\textbf{TRACE} (Temporal vAlidation with refinement and Consistency 
Estimation) adds multi-frame temporal consistency validation and 
two-phase backward FFBO refinement to SOLE, but omits motion scoring. Crucially, temporal persistence alone cannot separate the active tackler from a stationary bystander who appears consistently near the dummy, nor from a player still repositioning at drill onset who moves laterally rather than forward. Without a directional motion filter, TRACE rejects these valid players at the validation stage, collapsing coverage to 
46.9\%, confirming that temporal consistency \emph{requires} directional 
motion scoring to be discriminative rather than merely restrictive.

Baseline \textbf{B3: }\textbf{MARS} (Motion-Aware Region Selection) adds player motion scoring  to SOLE while omitting temporal validation and backward refinement. Its  coverage (91.9\%) and conditional precision (85.7\% at 
$\epsilon{=}20$) are nearly identical to SOLE (92.0\%, 85.8\%), showing that motion scoring in isolation changes candidate ranking upstream but does not improve per-clip localization accuracy. Its benefit emerges in the full system, where it interacts with multi-prompt progressive search and multi-candidate SAM2 retry to suppress distractors before they reach the segmentation stage.

\textbf{GRAZE} is the complete pipeline: hierarchical multi-prompt 
progressive search, temporal validation, directional motion scoring, two-phase backward refinement, and ranked multi-candidate SAM2 retry.
%
\section{Results}

\begin{figure}[t]
    \centering
    \includegraphics[width=\linewidth]{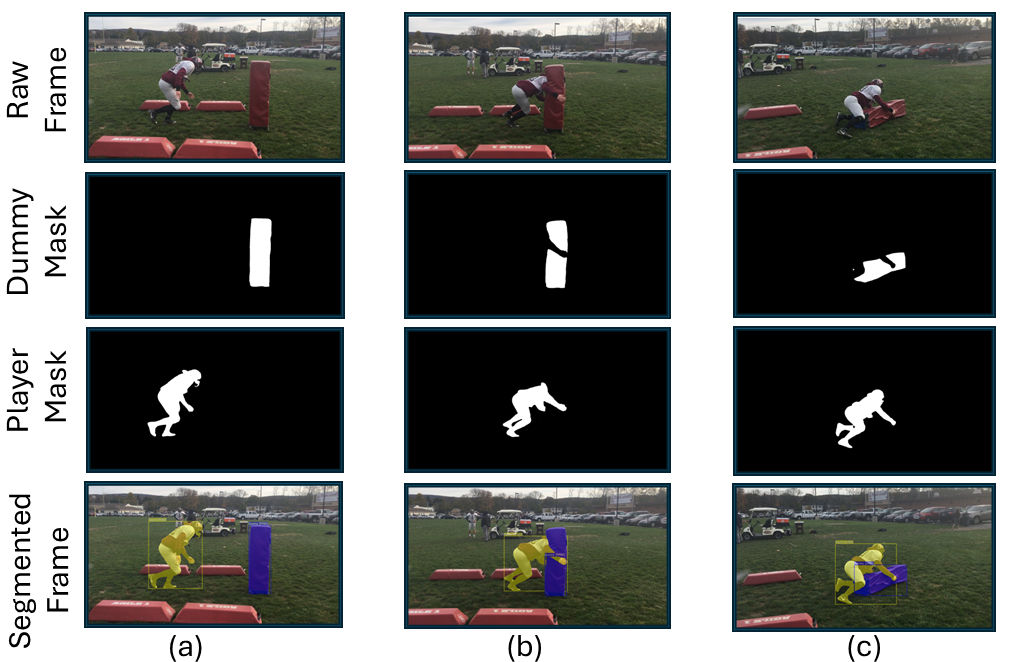}
    \caption{Representative tackle-event frames. Columns show
    (a)~$t_{\text{FFBO}}$ (first frame with both objects visible),
    (b)~$t_{\text{FPOC}}$ (contact onset), and (c)~a post-contact frame
    near $t_{\text{end}}$. Rows show the raw frame, the dummy mask,
    the player mask, and the composite overlay.}
    \label{fig:qualitative}
    \vspace{-0.8em}
\end{figure}

We evaluated GRAZE on 738 untrimmed practice videos, of which 681 carry frame-accurate ground-truth FPOC annotations. We report three metrics. \emph{Coverage} is the fraction of videos for which the pipeline produces any output. \emph{End-to-end accuracy} treats all 738 clips as the denominator, penalizing videos for which the system produces no output. \emph{Conditional precision} restricts the denominator to the evaluable subset, clips where both a system output and a ground-truth label exist, isolating per-clip localization quality from coverage effects. Of GRAZE's 719 outputs, 666 overlap with the 681 GT-labeled clips; the remaining 53 segmented clips lack annotations, and 15 GT-labeled clips fell into manual review. A representative segmentation is shown in Figure~\ref{fig:qualitative}.

\subsection{Coverage}
\label{sec:results_coverage}

\begin{figure*}[t]
    \centering
    \includegraphics[width=\linewidth]{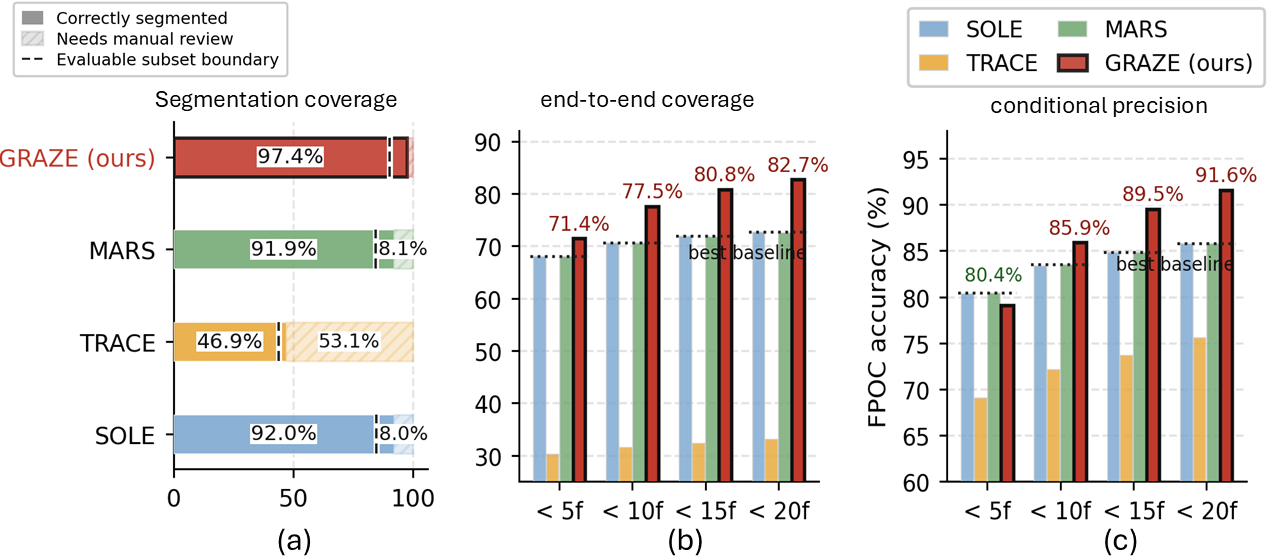}
    \caption{Ablation study across three performance axes.
    \textbf{(a)}~Segmentation coverage on all 738 videos. Solid bars
    show correctly segmented clips; hatched bars show clips routed to
    manual review; the dashed boundary marks the evaluable subset.
    \textbf{(b)}~End-to-end FPOC accuracy at four frame-error
    tolerances over all 738 clips. Dotted lines mark the best-baseline
    value per tolerance.
    \textbf{(c)}~Conditional FPOC precision over the evaluable subset.
    At $\pm5$ frames SOLE (80.3\%) and MARS (80.4\%) marginally lead
    GRAZE (79.1\%); the full-system advantage emerges and widens from
    $\pm10$ frames onward.}
    \label{fig:main}
    \vspace{-0.8em}
\end{figure*}

Coverage results appear in Figure~\ref{fig:main}(a) and the top rows
of Table~\ref{tab:ablation}. GRAZE produces valid outputs on 719 of
738 videos (97.4\%), routing only 19 clips to manual review.
SOLE (grounding and SAM2 only) and MARS (SOLE plus motion scoring)
each process around 92\% of the dataset; the roughly 40-clip gap
relative to GRAZE is closed by multi-prompt progressive search and
multi-candidate SAM2 retry, neither of which is present in either
baseline.

TRACE (SOLE plus temporal validation and backward refinement) is the
outlier at 46.9\%. The failure is not localization error but coverage
collapse. Without a directional motion gate, temporal validation
becomes unreliable in crowded scenes: detections can shift between
nearby athletes across validation frames, lowering match consistency
below the acceptance threshold and causing otherwise valid candidates
to be rejected. TRACE then routes these clips to manual review rather
than passing a candidate to SAM2. Directional scoring prevents this by
requiring candidates to show clear approach motion toward the dummy,
making the validation stage genuinely discriminative.

\subsection{Localization Accuracy}
\label{sec:results_accuracy}

Table~\ref{tab:ablation} summarizes results at all tolerance levels.
On the full 738-clip measure, GRAZE correctly localizes FPOC in 71.4\%
of videos within $\pm$5 frames, 77.5\% within $\pm$10 frames, and
82.7\% within $\pm$20 frames. Margins over SOLE and MARS grow with
tolerance, 3.4, 6.9, and 10.1 points at the three thresholds. Because
backward refinement primarily rescues clips where the initial grounding anchor is temporally displaced from the true onset, the benefit is most visible at wider tolerances.

Conditional precision at $\epsilon = 20$ frames reaches 91.6\% for
GRAZE, against 85.7-85.8\% for SOLE and MARS. The advantage grows
from 2.5 points at $\epsilon = 10$ frames to 5.8 points at
$\epsilon = 20$, consistent with refinement having its largest effect
on clips with the biggest temporal offset.

The one reversal is at $\epsilon < 5$ frames, where MARS reaches
80.4\% and SOLE 80.3\%, against GRAZE's 79.1\%. This is a genuine
trade-off: backward refinement occasionally steps one to two frames
past the true onset, shifting a small number of near-perfect
predictions into the adjacent error band. The payoff is a halved
catastrophic-error rate, GRAZE's $|\text{err}| \geq 20$ frame rate
is 8.4\%, versus 14.2\% and 14.3\% for SOLE and MARS. For downstream biomechanical measurement, a two-frame overshoot is far less damaging
than a twenty-frame miss, so the tail reduction is the more
consequential result.

\begin{table}[t]
\centering
\footnotesize
\setlength{\tabcolsep}{5pt}
\renewcommand{\arraystretch}{1.1}
\caption{FPOC localization results on 738 videos (681 with
ground-truth annotations). Frame error is
$|\hat{t}_{\text{FPOC}} - t^{*}_{\text{FPOC}}|$.
End-to-end accuracy uses all 738 clips as the denominator.
Conditional precision uses the evaluable subset (system
output $\cap$ GT-labeled pool).
$^\dagger$GRAZE segments 719 clips; 666 carry ground-truth
labels (53 segmented clips lack annotations; 15 GT-labeled
clips fell into manual review).}
\label{tab:ablation}
\begin{tabular}{lcccc}
\toprule
\textbf{Metric} & \textbf{SOLE} & \textbf{TRACE} & \textbf{MARS}
    & \textbf{GRAZE (ours)} \\
\midrule
\multicolumn{5}{l}{\textit{Coverage}}\\
Correctly segmented      & 679  & 346  & 678  & \textbf{719} \\
Coverage (\%)            & 92.0 & 46.9 & 91.9 & \textbf{97.4} \\
Evaluable clips$^\dagger$& 625  & 323  & 624  & \textbf{666} \\
\midrule
\multicolumn{5}{l}{\textit{End-to-end accuracy (\%): all 738 videos}}\\
$|\text{err}| < 5$\,f  & 68.0 & 30.2 & 68.0 & \textbf{71.4} \\
$|\text{err}| < 10$\,f & 70.6 & 31.6 & 70.6 & \textbf{77.5} \\
$|\text{err}| < 15$\,f & 71.8 & 32.2 & 71.7 & \textbf{80.8} \\
$|\text{err}| < 20$\,f & 72.6 & 33.1 & 72.5 & \textbf{82.7} \\
\midrule
\multicolumn{5}{l}{\textit{Conditional precision (\%): evaluable clips}}\\
$|\text{err}| < 5$\,f  & \textbf{80.3} & 69.0 & \textbf{80.4} & 79.1 \\
$|\text{err}| < 10$\,f & 83.4 & 72.1 & 83.5 & \textbf{85.9} \\
$|\text{err}| < 15$\,f & 84.8 & 73.7 & 84.8 & \textbf{89.5} \\
$|\text{err}| < 20$\,f & 85.8 & 75.5 & 85.7 & \textbf{91.6} \\
\midrule
\multicolumn{5}{l}{\textit{Error tail (conditional)}}\\
$|\text{err}| \ge 20$\,f (\%) & 14.2 & 24.5 & 14.3 & \textbf{8.4} \\
\bottomrule
\end{tabular}
\end{table}


Figure~\ref{fig:qualitative} shows a representative successful example
from the evaluable set. SAM2 maintains a stable dummy mask across the
contact window despite partial occlusion during impact, and the player
mask deforms through the sprint-to-contact transition without identity
loss. The predicted $t_{\text{FPOC}}$ corresponds to the earliest frame
of nonzero mask intersection, matching the annotator-defined onset.

The most common error in correctly processed clips is
approach-from-behind geometry: in the final stride, the player's
image-plane silhouette overlaps the dummy before physical contact
occurs, typically placing the prediction two to four frames early.
This directly accounts for the shortfall at $\epsilon < 5$ frames and
is a fundamental limitation of 2D mask overlap as a contact proxy.

Among the 19 unprocessed clips, two patterns account for nearly all
failures. Off-camera approaches produce no simultaneous player-dummy
view, making grounding impossible regardless of prompt or threshold.
In concurrent-player scenes two athletes approach within the same short
window; neither clears the directional motion threshold independently,
so both are discarded and the clip is flagged for review.

\section{Conclusion}

We presented GRAZE, a training-free pipeline for frame-accurate contact onset localization in unconstrained American football practice footage. The central contribution is a reframing of SAM2 mask propagation: rather than treating it as a passive segmentation backend, GRAZE uses it as an independent pixel-level contact verifier wholly decoupled from grounding confidence. This separation enables a ranked multi-candidate fallback strategy in which all hypotheses generated across prompt hierarchies, temporal positions, and detection thresholds are tried in quality order until mask intersection independently confirms contact.

Applied to 738 unconstrained practice videos without any domain-specific fine-tuning, GRAZE achieves 97.4\% coverage and conditional FPOC precision of 91.6\% within $\pm$20 frames and 85.9\% within $\pm$10 frames. The ablation isolates the contribution of each stage: multi-prompt progressive search closes the coverage gap from 92\% to 97.4\%; directional motion scoring is what makes temporal validation discriminative rather than merely restrictive, as removing it (TRACE)  collapses coverage to 46.9\% by discarding valid players who fail persistence tests designed for stationary objects; and two-phase backward refinement halves the catastrophic-error rate relative to the strongest single-component baselines SOLE and MARS.

The 2.6\% of unresolved clips expose two structural limits. Off-camera approaches produce no simultaneous player-dummy view, making grounding geometrically impossible regardless of prompt quality. Concurrent multi-player scenes expose the limits of per-candidate directional scoring, which has no mechanism to compare relative approach trajectories across athletes. Future work will address both through a temporal multi-object tracker that maintains identity continuity across frames, replacing per-frame grounding disambiguation. Replacing image-plane mask overlap with depth-aware contact verification is a natural extension that would tighten the FPOC criterion for approach-from-behind trajectories, while preserving the core deployment constraint of zero domain-specific annotation at any pipeline stage.

{\small
\bibliographystyle{ieeenat_fullname}
\bibliography{references}
}

\end{document}